\documentclass[10pt,twocolumn,letterpaper]{article}

\usepackage[pagenumbers]{cvpr}
\usepackage{multirow}

\usepackage{tikz}
\usetikzlibrary{arrows.meta,positioning,fit,calc}
\usepackage{pgfplots}
\pgfplotsset{compat=1.18}
\definecolor{vlmblue}{RGB}{42,105,155}
\definecolor{expertorange}{RGB}{190,100,34}

\definecolor{cvprblue}{rgb}{0.21,0.49,0.74}
\usepackage[pagebackref,breaklinks,colorlinks,allcolors=cvprblue]{hyperref}

\title{VLM-guided Grounding Expert for Generalised Referring Expression Comprehension}

\author{
Spirit AI Agent Team\thanks{
The full author list and affiliations are provided
in Appendix~\ref{sec:authors}.
}\\
Spirit AI, Beijing, China
}
\date{}

\begin{document}
\maketitle

\begin{abstract}
Generalised referring expression comprehension (GREC) aims to identify and localise the zero, one or multiple objects referred to by a language expression. In robotic manipulation, grounding also needs to remain accurate when requests for the same objects are paraphrased or embedded in additional context in robotics.
However, grounding expert models are sensitive to instruction variation, while generative vision--language models (VLMs) struggle with complete target recovery and rejection of absent targets. These complementary limitations motivate combining the contextual representations of VLMs with the direct spatial prediction capabilities of grounding experts. 
We therefore introduce VLM-guided Grounding Expert (VGE), which combines a frozen VLM with a trainable grounding expert through layerwise joint attention and text-conditioned instance scoring. These interfaces leverage multimodal understanding of the VLM to guide object-query refinement and candidate selection while retaining direct bounding-box prediction. 
Across three datasets, VGE achieves the highest averaged F1 score among the evaluated methods on all five test splits, with differences of at most 1.2 percentage points across instruction conditions within each split. On the dataset with robotic scenes, VGE also achieves the highest F1 across instruction variants and no-target, single-target and multiple-target cases. VGE provides a $5.3$--$13.0\times$ inference speedup over the baseline with the same VLM base model. These results support combining VLM context with specialised spatial prediction to address both grounding challenges, with particularly strong performance in robotic scenes. Code will be released upon publication.
\end{abstract}

\section{Introduction}
\label{sec:intro}

\begin{figure}[ht!]
  \centering
  \includegraphics[width=0.95\linewidth]{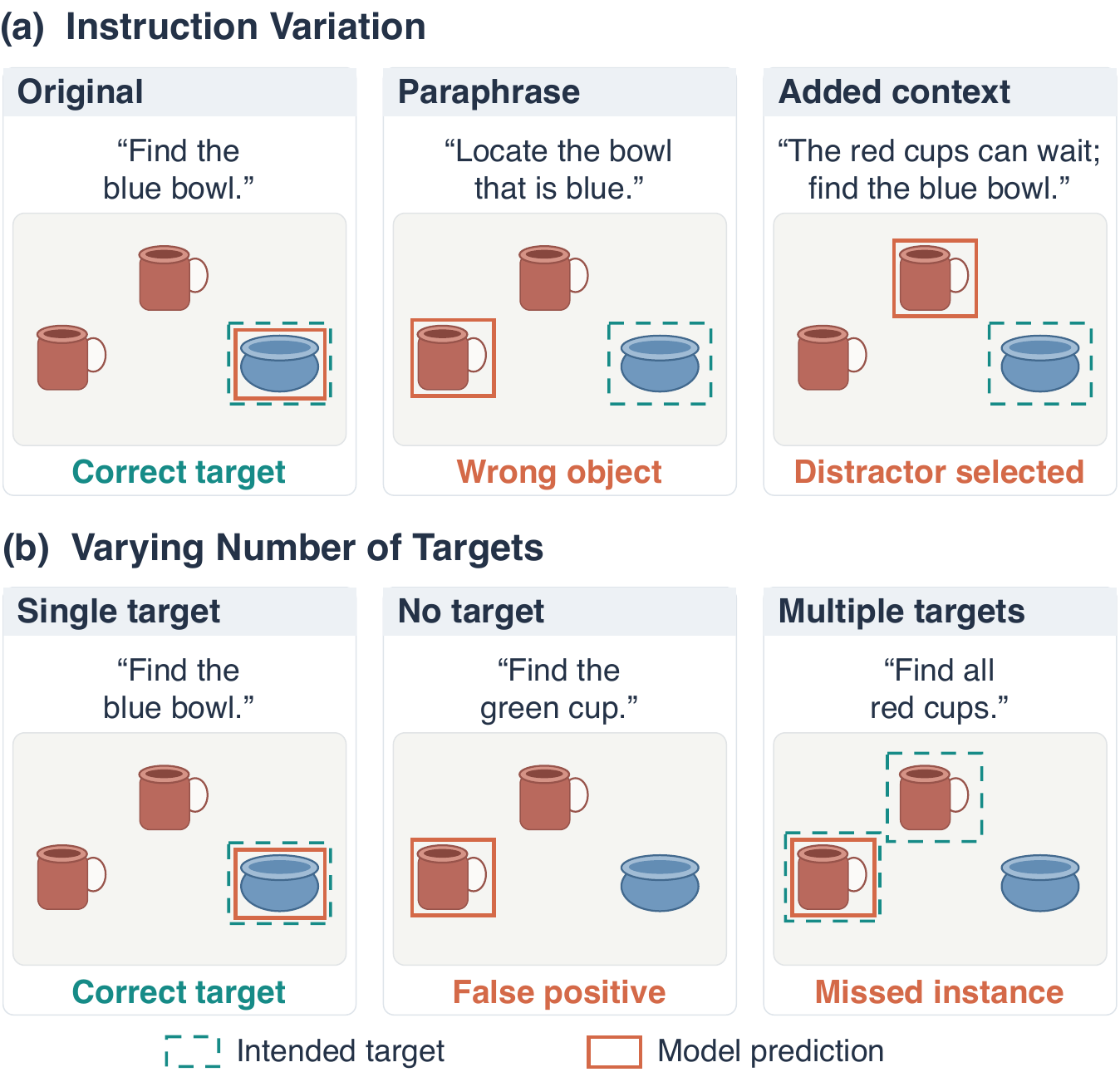}
  \caption{\textbf{Challenges in robotic grounding.} (a) Instruction variation can change predictions despite an unchanged target. (b) Target-set errors include selecting absent objects and missing requested instances. Dashed teal boxes mark intended targets; solid orange boxes mark illustrative predictions. All examples are schematic and use the same scene.}
\label{fig:challenges}

\end{figure}

Identifying and localising objects specified by language instructions is important for robotic tasks, including coordinated robot-arm manipulation~\cite{duan2026robopara} and long-horizon object manipulation~\cite{li2025objectcentric}. Spatial cues may also support action prediction in vision--language--action (VLA) models~\cite{wang2026vpvla}.  However, grounding language instructions in robotic scenes involves both varying numbers of targets and instruction variation. Generalised referring expression comprehension (GREC) addresses the former by extending conventional single-target grounding to expressions referring to zero, one or multiple objects~\cite{he2023grec,hemanthage2024recantformer}. Instruction variation arises when users express the same manipulation request through different action descriptions and object references~\cite{kim2026liberopara}. In human--robot interaction, requests also appear in longer contexts that include explanations or incidental remarks, adding irrelevant information without necessarily changing the intended targets~\cite{moskalenko2026bring}. Figure~\ref{fig:challenges} illustrates these two challenges. Reliable grounding therefore requires predicting a complete, potentially empty target set while maintaining correctness across semantically equivalent instructions.

Despite progress in visual grounding, the joint challenges of varying numbers of targets and instruction variation remain insufficiently explored. Grounding expert models such as Grounding DINO~\cite{liu2024groundingdino} learn image--text alignment with dedicated box-prediction modules~\cite{liu2024groundingdino}, while general vision--language models (VLMs), such as Qwen3.5~\cite{qwen2026qwen35} and Gemma4~\cite{team2026gemma}, generate object locations in text based on their multimodal understanding. Our diagnostic evaluation identifies distinct limitations in grounding experts and VLMs. Grounding experts remain sensitive to instruction variation, while VLMs struggle to localise all matching objects and recognise when no valid target exists. These complementary failure modes motivate combining the relative robustness of VLMs to instruction variation with the spatial prediction capabilities of grounding experts.

In this work, we introduce the VLM-guided Grounding Expert (VGE), which combines a frozen VLM with a trainable grounding expert to condition object localisation and target selection based on the multimodal understanding of the VLM. Since grounding requires both refining candidate representations and determining which instances match the instruction, we introduce VLM guidance at both stages. 
Specifically, VGE incorporates the VLM's contextual representations into the grounding expert through two interfaces. Layerwise joint attention conditions the expert’s object-query refinement on VLM context, allowing instruction information to guide spatial prediction within the expert. Text-conditioned instance scoring complements this process by assessing each candidate’s relevance to the instruction. 
By using the VLM as a frozen representation provider rather than an autoregressive coordinate generator, VGE retains direct prediction of the grounding expert, maintaining high prediction efficiency.
Together, these interfaces use instruction context to guide both box prediction and the selection of zero, one, or multiple targets, preserving native visual processing and direct box prediction. 
We make three contributions.
\begin{itemize}
    \item We conduct a diagnostic evaluation of grounding experts and VLMs under instruction variation and varying numbers of targets, identifying distinct limitations in instruction robustness and object localisation and selection.

    \item We propose VGE, connecting a frozen VLM to a trainable expert via layerwise joint attention and text-conditioned instance scoring. These interfaces guide target localisation and selection using multimodal context.

    \item Across three datasets, VGE achieves the highest averaged F1 among the evaluated methods on all five test splits. On RoboRefIt, it also leads across all evaluated cases while providing a $5.3$--$13.0\times$ inference speedup over Qwen3.5-9B across the three datasets. Training only the grounding expert and interfaces provides a competitive solution compared with additional VLM parameter updates.
\end{itemize}

\section{Related Work}
\label{sec:related}

\subsection{Visual Grounding}
\label{sec:related_grounding}

Visual grounding establishes correspondences between natural-language descriptions and regions in an image. A central task is referring expression comprehension (REC), which aims to identify and localise the object specified by an expression. 
Language-conditioned detectors can be used for this task. Grounding DINO integrates grounded pre-training, language-guided query selection and cross-modal fusion to locate objects described by category names or referring expressions~\cite{liu2024groundingdino}. YOLO-World learns region--text alignment through contrastive learning~\cite{cheng2024yoloworld}, while LLMDet introduces language-model supervision during detector training and removes the language model at inference~\cite{fu2025llmdet}. GREC extends REC to expressions referring to zero, one or multiple objects~\cite{he2023grec}. RECANTFormer addresses varying target counts with an encoder-only transformer and Hungarian matching between predicted and ground-truth boxes~\cite{hemanthage2024recantformer}. 
However, our evaluation shows that these methods are sensitive to added context even when the intended targets are unchanged.

Generative VLMs perform visual grounding through multimodal instruction following. Shikra represents spatial coordinates as numerical text~\cite{chen2023shikra}, while Qwen3.5 can generate bounding-box coordinates from image--text inputs~\cite{qwen2026qwen35}. This formulation allows the language model to use multimodal context when interpreting a request and producing spatial outputs. However, language understanding and grounding accuracy need to be assessed separately. Ref-Adv shows that strong conventional REC performance can coexist with difficulties resolving expressions that require distinguishing targets from challenging visual distractors~\cite{dong2026refadv}. Autoregressive coordinate generation also incurs sequential decoding overhead. VGent reports increasing inference time with the number of output boxes for the evaluated generative VLMs~\cite{kang2026vgent}. ECG Mirage also shows that strong predictive performance can coexist with limited use of visual evidence~\cite{liang2026ecg}. These findings motivate combining VLM-based instruction understanding with dedicated localisation modules that predict boxes directly.

\subsection{Integrating VLMs with Localisation Modules}
\label{sec:related_integration}

Existing VLM baselines connect multimodal representations to spatial predictions through generated tokens, attention maps or task-specific modules. In segmentation, LISA passes the hidden representation of a generated segmentation token to a mask decoder~\cite{lai2024lisa}. F-LMM instead converts attention maps from a frozen multimodal model into masks and refines them with a SAM-based module~\cite{wu2025flmm}.  VGE also keeps the VLM frozen, but provides fine-grained multimodal context with layerwise joint attention and object-query refinement with instance scoring. 

Other methods extract regions of interest (RoI) first and input them into VLMs. Groma encodes RoI as visual tokens for region-level understanding and grounded responses~\cite{ma2024groma}. ROD-MLLM associates detected regions with region-anchor tokens that the language model generates to identify referred objects, supporting expressions with no matching target or multiple matching targets~\cite{yin2025rodmllm}. These methods incorporate region-specific representations into language-model processing, yet they still rely on the VLM's linguistic capabilities. In contrast, VGE enables spatial understanding by leveraging both the VLM’s multimodal understanding and specialised grounding experts.

\section{Methodology}
\label{sec:method}

\subsection{Problem Formulation}
\label{sec:problem}

Given an image $I$ and a natural-language instruction $P$, instruction-conditioned object grounding aims to identify and localise all object instances that satisfy the request. A grounding model $F$ predicts an unordered set of bounding boxes $\widehat{\mathcal B}=F(I,P)$. The corresponding ground-truth set is
\begin{equation}
    \mathcal B^*(I,P)=\{b_j\}_{j=1}^{M},
    \label{eq:grounding}
\end{equation}
where $b_j\in[0,1]^4$ denotes the normalised coordinates of the $j$-th target box and $M$ is the number of matching instances.

The challenge of varying numbers of targets arises because the number of matching objects depends on both the image and the instruction. We consider three cases,
\begin{equation}
M=|\mathcal B^*(I,P)|, \ 
M=\begin{cases}
{0}, & \text{no-target},\\
{1}, & \text{single-target},\\
{>1}, & \text{multiple-target}.
\end{cases}
\label{eq}
\end{equation}
Across these cases, the model must localise all matching instances while excluding irrelevant candidates. When $M=0$, both sets should be empty.

Instruction variation concerns changes in wording or context that preserve the intended targets. For a given image $I$, let $T$ denote a target-preserving instruction transformation satisfying
\begin{equation}
    \mathcal B^*(I,T(P))=\mathcal B^*(I,P).
    \label{eq:target_preservation}
\end{equation}
We consider equivalent paraphrases and context-augmented instructions that leave the request unchanged. The model should recover the same matching objects from $P$ and $T(P)$, with accurate localisation under both instructions. Consistency does not imply correctness: unchanged predictions can miss targets or select distractors.

\subsection{VGE Architecture}
\label{sec:architecture}

\begin{figure*}[t]
 \centering
    \includegraphics[width=0.9\linewidth]{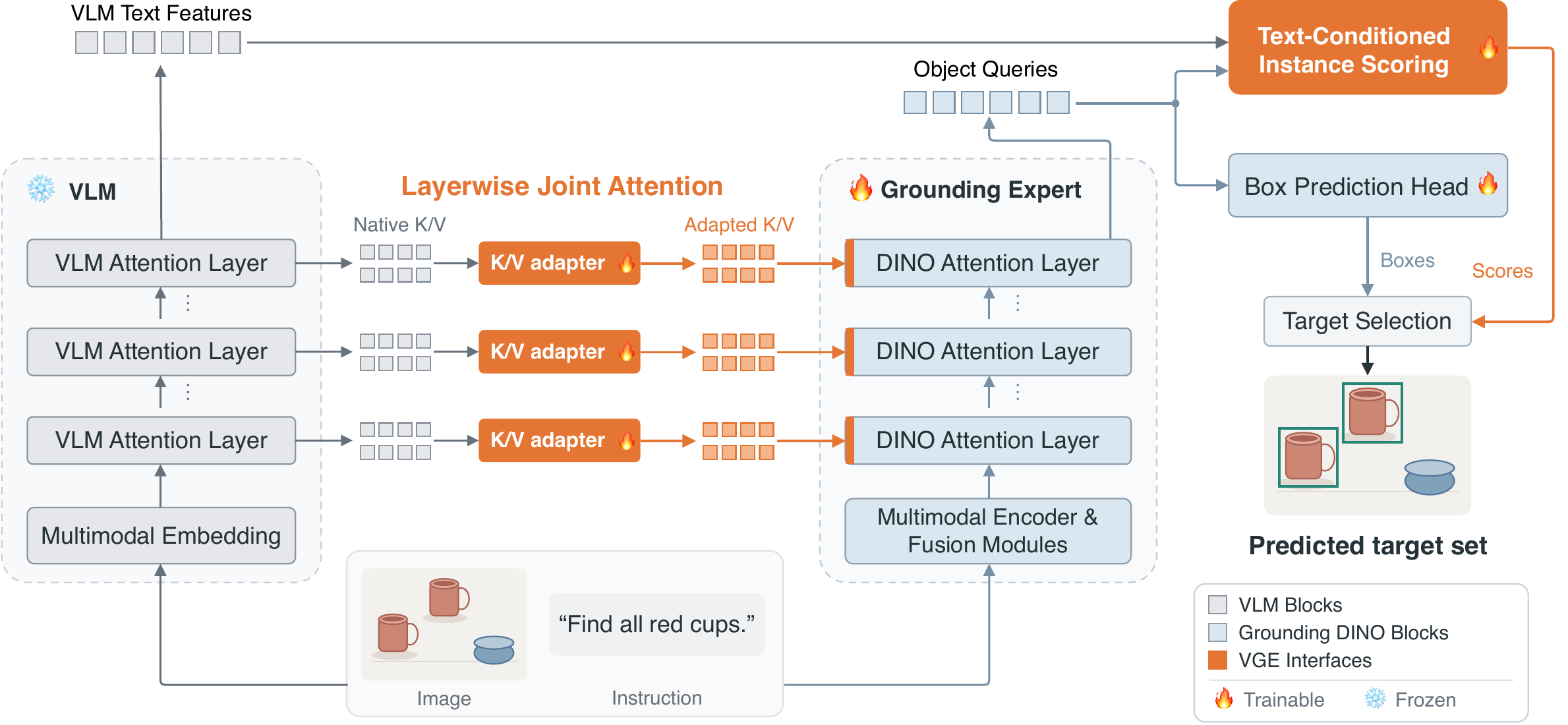}
    \vspace{-5px}
    \caption{\textbf{VGE architecture.} A frozen VLM supplies layerwise joint attention and contextualised text features to a trainable grounding expert. Gated joint attention refines object queries, whilst text-conditioned scoring adjusts native matching logits. The box prediction head produces candidate boxes; target selection combines boxes and scores into the predicted set.}
    \label{fig:architecture}
\end{figure*}

VGE instantiates the grounding model $F$ by combining a frozen VLM with a trainable grounding expert, as illustrated in Figure~\ref{fig:architecture}. Given an image $I$ and an instruction $P$, both models process the inputs using their respective preprocessing pipelines. The VLM provides contextual representations of the image and instruction, while the grounding expert performs object localisation and target selection. We use Grounding DINO~\cite{liu2024groundingdino} as the expert, retaining its visual backbone, text encoder, fusion encoder and detection decoder. 
We introduce two trainable interfaces to incorporate VLM context into complementary stages of grounding: object-query refinement and candidate scoring. Layerwise joint attention guides the former, while text-conditioned instance scoring adjusts the latter.
The expert's box-prediction head produces candidate bounding boxes, and confidence filtering followed by non-maximum suppression (NMS) yields the final predictions $\widehat{\mathcal B}$, which may contain zero, one, or multiple boxes. We train the complete grounding expert and both interfaces while keeping the VLM frozen, using its context to guide grounding without VLM-generated coordinates.

\subsection{Layerwise Joint Attention}
\label{sec:kv}

Layerwise joint attention allows the grounding expert's object queries to access multimodal understanding of the VLM. We pair the expert's six decoder layers, in depth order, with the VLM's last six full-attention layers. At decoder layer $\ell$, $Q_D^\ell$, $K_D^\ell$ and $V_D^\ell$ denote the expert's native self-attention queries, keys and values, while $K_L^\ell$ and $V_L^\ell$ denote the keys and values from the paired VLM layer. Trainable linear mappings $f_K^\ell$ and $f_V^\ell$ project the VLM keys and values into the expert's attention space,
\begin{equation}
    \widetilde K_L^\ell=f_K^\ell(K_L^\ell),
    \qquad
    \widetilde V_L^\ell=f_V^\ell(V_L^\ell),
    \label{eq:adapt}
\end{equation}
where $\widetilde K_L^\ell$ and $\widetilde V_L^\ell$ denote the adapted keys and values.

The expert queries attend jointly to the native expert features and adapted VLM features. We compute both the joint attention output and the native attention output,
\begin{equation}
    \begin{aligned}
        U_{\mathrm{joint}}^\ell
        &=\operatorname{Attn}\!\left(
        Q_D^\ell,
        K_D^\ell \bigoplus \widetilde K_L^\ell,
        V_D^\ell \bigoplus \widetilde V_L^\ell
        \right),\\
        U_{\mathrm{nat}}^\ell
        &=\operatorname{Attn}\!\left(
        Q_D^\ell,K_D^\ell,V_D^\ell
        \right),
    \end{aligned}
    \label{eq:joint}
\end{equation}
where $\bigoplus$ denotes concatenation along the token dimension and $\operatorname{Attn}$ is multi-head scaled dot-product attention~\cite{vaswani2017attention}, including a shared expert output projection. Joint attention uses a single softmax per head over the concatenated keys, so object queries can attend to expert and VLM features. Unlike the cross-attention operation, joint attention allows native expert features and projected VLM features to compete within the same attention distribution.

To preserve the expert's attention output at initialisation, we introduce a learnable scalar gate $\alpha_\ell$ at each layer,
\begin{equation}
    U^\ell=U_{\mathrm{nat}}^\ell+
    \alpha_\ell\left(
    U_{\mathrm{joint}}^\ell-U_{\mathrm{nat}}^\ell
    \right),
    \qquad
    \alpha_\ell\big|_{\mathrm{init}}=0,
    \label{eq:gate}
\end{equation}
where $U^\ell$ is the resulting attention output. Each gate is learned independently. We replace the expert's native self-attention output with $U^\ell$, retaining the subsequent residual, normalisation, and cross-attention operations. Supplementary \cref{sec:supp_interfaces} provides more implementation details.

\subsection{Text-conditioned Instance Scoring}
\label{sec:score}

While layerwise joint attention modifies the object-query representations used for prediction, it does not explicitly provide a separate correction to the final candidate scores. We therefore introduce text-conditioned instance scoring to adjust each candidate's relevance logit using contextualised VLM features.
Let $H_P$ denote the final VLM features at valid input text positions, contextualised by both the image $I$ and instruction $P$. A trainable linear mapping $f_H$, with parameters $\theta_H$, projects these features into the expert's object-query space. Each final object query $q_i$, after layer normalisation, attends to the projected features to obtain context $c_i$, which is used to correct its native matching logit,
\begin{equation}
    \begin{aligned}
        \widetilde H_P&=f_H(H_P;\theta_H),\\
        c_i&=\operatorname{Attn}\!\left(
        q_i,\widetilde H_P,\widetilde H_P
        \right),\\
        s_i&=s_i^{\mathrm{native}}+
        f_s\!\left([q_i;c_i]\right),
    \end{aligned}
    \label{eq:score}
\end{equation}
where $s_i$ is the corrected relevance logit for candidate $i$, and $s_i^{\mathrm{native}}$ is its native matching logit from the grounding expert. The trainable linear mapping $f_s$ maps the concatenated object query $q_i$ and context vector $c_i$ to a scalar score correction. Zero-initialising the weights of $f_s$ preserves the native scores at initialisation. The corrected logits $s_i$ determine which predicted boxes are retained through confidence filtering and NMS, yielding a potentially empty target set $\widehat{\mathcal B}$.

\subsection{Training Objective}
\label{sec:training}

We train the grounding expert and both interfaces while keeping the VLM frozen, allowing the expert to learn from fixed multimodal representations without updating the VLM parameters. Supervision is applied only to the expert's final-layer predictions. Hungarian matching assigns ground-truth boxes in $\mathcal B^*(I,P)$ to predicted candidates~\cite{carion2020detr}, and the training objective is
\begin{equation}
    \mathcal L
    =\mathcal L_{\mathrm{focal}}
    +\lambda_1\mathcal L_{\ell_1}
    +\lambda_2\mathcal L_{\mathrm{GIoU}},
    \label{eq:loss}
\end{equation}
where focal loss $\mathcal L_{\mathrm{focal}}$ supervises the corrected instance logits $s_i$~\cite{lin2017focal}, while $\mathcal L_{\ell_1}$ and $\mathcal L_{\mathrm{GIoU}}$ supervise matched bounding boxes~\cite{rezatofighi2019giou}. The coefficients $\lambda_1$ and $\lambda_2$ balance the regression terms. Unmatched queries receive negative classification targets. For no-target samples ($M=0$), all queries are negative, and both regression terms are zero. Supplementary \cref{sec:supp_interfaces} provides further training details.

\section{Results}
\label{sec:results}

We evaluate VGE through five research questions.
\begin{itemize}
    \item \textbf{RQ1:} How do instruction variation and varying numbers of targets affect grounding experts and VLMs?
    \item \textbf{RQ2:} How does VGE compare with baselines across instruction variants and varying numbers of targets?
    \item \textbf{RQ3:} Can training only the expert and interfaces match configurations that also adapt the VLM?
    \item \textbf{RQ4:} How does VGE compare with grounding experts and VLMs in inference latency?
    \item \textbf{RQ5:} How do layerwise joint attention and text-conditioned instance scoring contribute?
\end{itemize}

\subsection{Experimental Setup}
\label{sec:setup}

We evaluate on gRefCOCO~\cite{liu2023gres,he2023grec} and Ref-ZOM~\cite{Hu_2023_ICCV} for general scenes, and RoboRefIt~\cite{lu2023vlgrasp} for robotic scenes. We retain the original images, target annotations and test splits. Where no validation split is available, we reserve 1,200 proportionally sampled training references for validation. Each evaluation reference comprises an original instruction, a generated paraphrase and a context-augmented variant, all referring to the same annotated targets in the same image. Supplementary \cref{sec:supp_setup} describes the data splits and instruction-generation procedure.

Our main VGE configuration combines a frozen Qwen3.5-9B~\cite{qwen2026qwen35} with Grounding DINO-base and both proposed interfaces. All task-specific training uses only original instructions. RQ1 evaluates Qwen3.5-4B/9B, Grounding DINO-base and LLMDet-base~\cite{fu2025llmdet} without task-specific parameter updates. RQ2 compares VGE with Grounding DINO, LLMDet, RECANTFormer, and both zero-shot and LoRA-adapted versions of Qwen3.5-9B and Groma. For models that use confidence filtering, thresholds are calibrated on original validation instructions and held fixed across test splits and instruction variants within each dataset. Supplementary \cref{sec:supp_implementation} provides model configurations and training and inference details.

Following GREC~\cite{he2023grec}, we use an intersection-over-union (IoU) threshold of $0.5$ to determine valid box matches. We implement greedy one-to-one matching between predicted and ground-truth boxes. Exact-set accuracy requires every target to be matched without additional predictions. We also report mean sample F1, which averages per-reference F1 scores to capture partial recovery. For no-target references, valid empty predictions score one and non-empty predictions score zero~\cite{he2023grec}, so mean sample F1 equals no-target accuracy. Malformed or truncated outputs score zero and remain included in all comparisons for VLM baselines. Following prior work~\cite{zhu2026tokenpair}, we report standard deviations estimated from 1,000 bootstrap resamples. We resample references with pairing across methods and instruction variants, using predictions from a single evaluation run with seed~0.

\subsection{RQ1: Instruction Variation and Target Count}
\label{sec:rq1}

\begin{figure}[!htb]
\centering
\includegraphics[width=0.99\linewidth]{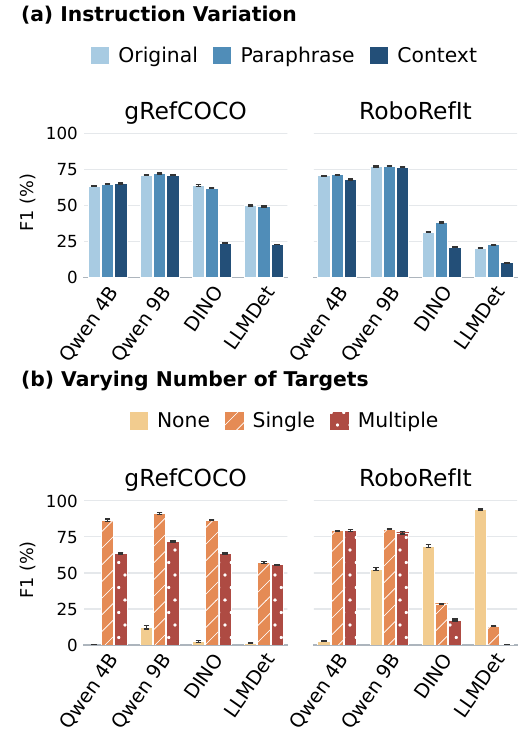}

\caption{\textbf{Grounding without task-specific fine-tuning.} Mean F1 (\%) on the Test A splits of gRefCOCO and RoboRefIt. (a) Original, paraphrased and context-augmented instructions. (b) No-target, single-target and multiple-target examples under original instructions. Qwen denotes Qwen3.5. Detector thresholds are determined on the validation set for each dataset. Error bars show one bootstrapped standard deviation.}
\label{fig:rq1}
\end{figure}

\begin{table*}[ht!]
\centering
\caption{\textbf{Grounding across instruction variants and varying numbers of targets.} Mean sample F1 (\%) on three datasets. Subscripts report bootstrap standard deviations from 1,000 reference-level resamples. Bold and underlined values indicate the highest and second-highest displayed scores in each column, respectively, with ties receiving the same formatting.}
\label{tab:rq2}
\begin{subtable}{\textwidth}
\centering
\caption{\textbf{Instruction variation.} Performance under original (Orig.), paraphrased (Para.) and context-augmented (Context) instructions.}
\scriptsize
\setlength{\tabcolsep}{1.8pt}
\renewcommand{\arraystretch}{1.30}
\resizebox{\linewidth}{!}{%
\begin{tabular}{@{}l*{15}{c}@{}}
\toprule
 & \multicolumn{6}{c}{gRefCOCO} & \multicolumn{6}{c}{RoboRefIt} & \multicolumn{3}{c}{Ref-ZOM} \\
\cmidrule(lr){2-7}\cmidrule(lr){8-13}\cmidrule(lr){14-16}
 & \multicolumn{3}{c}{Test A} & \multicolumn{3}{c}{Test B} & \multicolumn{3}{c}{Test A} & \multicolumn{3}{c}{Test B} & \multicolumn{3}{c}{Test} \\
\cmidrule(lr){2-4}\cmidrule(lr){5-7}\cmidrule(lr){8-10}\cmidrule(lr){11-13}\cmidrule(lr){14-16}
Method & Orig. & Para. & Context & Orig. & Para. & Context & Orig. & Para. & Context & Orig. & Para. & Context & Orig. & Para. & Context \\
\midrule
Grounding DINO & $73.0_{\pm 0.5}$ & $70.7_{\pm 0.5}$ & $63.5_{\pm 0.5}$ & $64.1_{\pm 0.6}$ & $62.3_{\pm 0.7}$ & $54.7_{\pm 0.7}$ & $\underline{94.4}_{\pm 0.2}$ & $89.8_{\pm 0.2}$ & $62.6_{\pm 0.3}$ & $\underline{89.7}_{\pm 0.3}$ & $83.9_{\pm 0.3}$ & $55.3_{\pm 0.3}$ & $83.6_{\pm 0.2}$ & $80.6_{\pm 0.2}$ & $75.4_{\pm 0.2}$ \\
LLMDet & $76.6_{\pm 0.5}$ & $72.5_{\pm 0.5}$ & $50.3_{\pm 0.4}$ & $68.5_{\pm 0.6}$ & $63.9_{\pm 0.6}$ & $44.9_{\pm 0.5}$ & $90.9_{\pm 0.2}$ & $87.8_{\pm 0.2}$ & $48.2_{\pm 0.4}$ & $87.6_{\pm 0.3}$ & $84.5_{\pm 0.3}$ & $53.9_{\pm 0.5}$ & $83.4_{\pm 0.2}$ & $81.3_{\pm 0.2}$ & $61.6_{\pm 0.2}$ \\
RECANTFormer & $45.4_{\pm 0.5}$ & $41.7_{\pm 0.5}$ & $33.4_{\pm 0.4}$ & $42.3_{\pm 0.7}$ & $38.4_{\pm 0.6}$ & $33.2_{\pm 0.5}$ & $44.3_{\pm 0.4}$ & $42.3_{\pm 0.4}$ & $39.1_{\pm 0.3}$ & $31.7_{\pm 0.4}$ & $29.8_{\pm 0.4}$ & $28.1_{\pm 0.4}$ & $43.5_{\pm 0.3}$ & $36.2_{\pm 0.3}$ & $37.2_{\pm 0.3}$ \\
\midrule
Qwen3.5-9B(Zero-shot) & $71.1_{\pm 0.5}$ & $72.0_{\pm 0.4}$ & $\underline{70.9}_{\pm 0.4}$ & $66.0_{\pm 0.6}$ & $67.4_{\pm 0.6}$ & $67.1_{\pm 0.6}$ & $77.0_{\pm 0.3}$ & $77.2_{\pm 0.3}$ & $76.4_{\pm 0.3}$ & $78.9_{\pm 0.4}$ & $79.8_{\pm 0.4}$ & $78.5_{\pm 0.4}$ & $79.2_{\pm 0.2}$ & $78.5_{\pm 0.2}$ & $79.4_{\pm 0.2}$ \\
Qwen3.5-9B(LoRA) & $\underline{79.2}_{\pm 0.5}$ & $\underline{74.1}_{\pm 0.5}$ & $68.2_{\pm 0.6}$ & $\mathbf{76.6}_{\pm 0.6}$ & $\mathbf{73.6}_{\pm 0.6}$ & $\underline{71.7}_{\pm 0.6}$ & $92.6_{\pm 0.2}$ & $\underline{91.5}_{\pm 0.2}$ & $\underline{92.5}_{\pm 0.2}$ & $87.6_{\pm 0.3}$ & $\underline{86.6}_{\pm 0.3}$ & $\underline{87.5}_{\pm 0.3}$ & $\mathbf{90.7}_{\pm 0.2}$ & $\mathbf{86.7}_{\pm 0.2}$ & $81.7_{\pm 0.2}$ \\
Groma (zero-shot) & $52.6_{\pm 0.4}$ & $52.9_{\pm 0.4}$ & $53.1_{\pm 0.4}$ & $53.0_{\pm 0.6}$ & $53.5_{\pm 0.6}$ & $52.1_{\pm 0.6}$ & $47.0_{\pm 0.3}$ & $47.3_{\pm 0.3}$ & $41.0_{\pm 0.3}$ & $40.4_{\pm 0.4}$ & $40.8_{\pm 0.4}$ & $32.1_{\pm 0.4}$ & $60.4_{\pm 0.2}$ & $59.4_{\pm 0.3}$ & $58.3_{\pm 0.3}$ \\
Groma (LoRA) & $72.6_{\pm 0.5}$ & $67.8_{\pm 0.5}$ & $70.5_{\pm 0.5}$ & $72.2_{\pm 0.6}$ & $68.4_{\pm 0.6}$ & $70.0_{\pm 0.6}$ & $92.1_{\pm 0.2}$ & $90.7_{\pm 0.2}$ & $91.4_{\pm 0.2}$ & $78.9_{\pm 0.4}$ & $77.0_{\pm 0.4}$ & $77.7_{\pm 0.4}$ & $\underline{86.1}_{\pm 0.2}$ & $80.1_{\pm 0.2}$ & $\underline{84.2}_{\pm 0.2}$ \\
\midrule
VGE (Ours) & $\mathbf{80.6}_{\pm 0.4}$ & $\mathbf{79.8}_{\pm 0.4}$ & $\mathbf{79.6}_{\pm 0.4}$ & $\underline{72.8}_{\pm 0.6}$ & $\underline{72.5}_{\pm 0.6}$ & $\mathbf{72.2}_{\pm 0.6}$ & $\mathbf{95.2}_{\pm 0.1}$ & $\mathbf{94.7}_{\pm 0.2}$ & $\mathbf{95.2}_{\pm 0.1}$ & $\mathbf{90.1}_{\pm 0.3}$ & $\mathbf{88.9}_{\pm 0.3}$ & $\mathbf{90.0}_{\pm 0.3}$ & $\underline{86.1}_{\pm 0.2}$ & $\underline{85.2}_{\pm 0.2}$ & $\mathbf{86.1}_{\pm 0.2}$ \\
\bottomrule
\end{tabular}%
}
\end{subtable}
\par\medskip
\begin{subtable}{\textwidth}
\centering
\caption{\textbf{Varying numbers of targets.} Performance on no-, single-, and multiple-target cases, averaged equally over the three instruction conditions.}
\scriptsize
\setlength{\tabcolsep}{1.8pt}
\renewcommand{\arraystretch}{1.30}
\resizebox{\linewidth}{!}{%
\begin{tabular}{@{}l*{15}{c}@{}}
\toprule
 & \multicolumn{6}{c}{gRefCOCO} & \multicolumn{6}{c}{RoboRefIt} & \multicolumn{3}{c}{Ref-ZOM} \\
\cmidrule(lr){2-7}\cmidrule(lr){8-13}\cmidrule(lr){14-16}
 & \multicolumn{3}{c}{Test A} & \multicolumn{3}{c}{Test B} & \multicolumn{3}{c}{Test A} & \multicolumn{3}{c}{Test B} & \multicolumn{3}{c}{Test} \\
\cmidrule(lr){2-4}\cmidrule(lr){5-7}\cmidrule(lr){8-10}\cmidrule(lr){11-13}\cmidrule(lr){14-16}
Method & None & Single & Multiple & None & Single & Multiple & None & Single & Multiple & None & Single & Multiple & None & Single & Multiple \\
\midrule
Grounding DINO & $62.7_{\pm 1.6}$ & $76.8_{\pm 0.8}$ & $66.1_{\pm 0.5}$ & $59.5_{\pm 1.6}$ & $63.2_{\pm 1.0}$ & $58.1_{\pm 0.7}$ & $70.5_{\pm 0.5}$ & $84.1_{\pm 0.2}$ & $81.6_{\pm 0.4}$ & $65.2_{\pm 0.7}$ & $78.5_{\pm 0.3}$ & $69.5_{\pm 0.8}$ & $85.7_{\pm 0.6}$ & $80.7_{\pm 0.3}$ & $76.7_{\pm 0.3}$ \\
LLMDet & $51.9_{\pm 0.9}$ & $66.5_{\pm 0.6}$ & $69.4_{\pm 0.4}$ & $50.2_{\pm 1.0}$ & $54.9_{\pm 0.8}$ & $65.9_{\pm 0.6}$ & $65.9_{\pm 0.5}$ & $77.6_{\pm 0.3}$ & $72.0_{\pm 0.4}$ & $60.2_{\pm 0.6}$ & $77.9_{\pm 0.3}$ & $69.2_{\pm 0.6}$ & $83.1_{\pm 0.5}$ & $74.9_{\pm 0.2}$ & $74.0_{\pm 0.3}$ \\
RECANTFormer & $40.3_{\pm 1.3}$ & $36.6_{\pm 0.8}$ & $42.2_{\pm 0.5}$ & $40.3_{\pm 1.3}$ & $34.3_{\pm 0.9}$ & $40.5_{\pm 0.7}$ & $61.7_{\pm 0.9}$ & $38.7_{\pm 0.4}$ & $44.1_{\pm 0.7}$ & $61.5_{\pm 1.2}$ & $26.0_{\pm 0.4}$ & $28.0_{\pm 1.0}$ & $82.9_{\pm 0.7}$ & $25.9_{\pm 0.3}$ & $46.4_{\pm 0.4}$ \\
\midrule
Qwen3.5-9B(Zero-shot) & $27.2_{\pm 1.2}$ & $\mathbf{90.4}_{\pm 0.6}$ & $69.7_{\pm 0.4}$ & $14.8_{\pm 0.9}$ & $\mathbf{83.3}_{\pm 0.8}$ & $\underline{70.0}_{\pm 0.6}$ & $63.0_{\pm 1.0}$ & $79.0_{\pm 0.3}$ & $76.4_{\pm 0.7}$ & $64.8_{\pm 1.3}$ & $81.4_{\pm 0.4}$ & $74.4_{\pm 1.1}$ & $64.9_{\pm 0.7}$ & $\mathbf{86.2}_{\pm 0.3}$ & $71.8_{\pm 0.3}$ \\
Qwen3.5-9B(LoRA) & $\mathbf{89.4}_{\pm 0.9}$ & $73.9_{\pm 0.7}$ & $\underline{70.5}_{\pm 0.6}$ & $\mathbf{83.8}_{\pm 1.1}$ & $\underline{75.1}_{\pm 0.9}$ & $69.6_{\pm 0.8}$ & $96.1_{\pm 0.3}$ & $\underline{91.9}_{\pm 0.2}$ & $\underline{90.7}_{\pm 0.4}$ & $\underline{94.7}_{\pm 0.5}$ & $\underline{87.6}_{\pm 0.3}$ & $\underline{76.3}_{\pm 1.0}$ & $\mathbf{99.2}_{\pm 0.2}$ & $83.8_{\pm 0.3}$ & $\mathbf{86.6}_{\pm 0.2}$ \\
Groma (zero-shot) & $0.0_{\pm 0.0}$ & $58.6_{\pm 0.6}$ & $60.5_{\pm 0.4}$ & $0.0_{\pm 0.0}$ & $63.1_{\pm 0.7}$ & $61.9_{\pm 0.6}$ & $0.0_{\pm 0.0}$ & $49.2_{\pm 0.3}$ & $60.9_{\pm 0.8}$ & $0.0_{\pm 0.0}$ & $41.4_{\pm 0.4}$ & $47.7_{\pm 1.1}$ & $0.0_{\pm 0.0}$ & $66.7_{\pm 0.3}$ & $66.1_{\pm 0.3}$ \\
Groma (LoRA) & $58.2_{\pm 1.5}$ & $80.5_{\pm 0.8}$ & $67.0_{\pm 0.6}$ & $74.7_{\pm 1.4}$ & $74.4_{\pm 0.9}$ & $64.9_{\pm 0.7}$ & $\underline{96.3}_{\pm 0.3}$ & $90.9_{\pm 0.2}$ & $90.1_{\pm 0.4}$ & $94.0_{\pm 0.5}$ & $77.5_{\pm 0.4}$ & $64.0_{\pm 1.0}$ & $97.0_{\pm 0.3}$ & $84.4_{\pm 0.3}$ & $77.7_{\pm 0.3}$ \\
\midrule
VGE (Ours) & $\underline{86.0}_{\pm 1.2}$ & $\underline{84.0}_{\pm 0.8}$ & $\mathbf{76.5}_{\pm 0.5}$ & $\underline{75.7}_{\pm 1.4}$ & $73.0_{\pm 1.0}$ & $\mathbf{71.0}_{\pm 0.7}$ & $\mathbf{97.5}_{\pm 0.3}$ & $\mathbf{94.5}_{\pm 0.2}$ & $\mathbf{96.0}_{\pm 0.3}$ & $\mathbf{96.2}_{\pm 0.5}$ & $\mathbf{89.8}_{\pm 0.3}$ & $\mathbf{81.3}_{\pm 1.0}$ & $\underline{97.7}_{\pm 0.3}$ & $\underline{84.5}_{\pm 0.3}$ & $\underline{84.1}_{\pm 0.3}$ \\
\bottomrule
\end{tabular}%
}
\end{subtable}
\par\smallskip
\end{table*}

In this experiment, we examine how instruction variation and varying numbers of targets affect grounding experts and VLMs without task-specific training. We evaluate Qwen3.5-4B/9B, Grounding DINO-base and LLMDet-base on the Test A splits of gRefCOCO and RoboRefIt. 

As shown in Figure~\ref{fig:rq1}(a), Grounding DINO and LLMDet lose performance under context augmentation on both datasets, whereas paraphrasing has smaller and sometimes positive effects. Their sensitivity therefore depends on how the instruction changes, with additional context presenting a greater challenge than alternative wording. Our comparison further shows that performance can decline even when the image and intended targets remain unchanged. On RoboRefIt, both grounding experts also perform better on no-target than target-containing examples under original instructions (Figure~\ref{fig:rq1}(b)), indicating that successful rejection does not necessarily accompany accurate target localisation.

VLM baselines like Qwen3.5-4B and Qwen3.5-9B show comparatively stable performance across instruction variants, but both perform substantially worse on no-target than single-target samples. This distinction highlights that robustness at the aggregate level does not establish reliable identification of whether a requested object is present. Related work on object hallucination likewise motivates evaluating object presence explicitly in VLMs~\cite{li2023evaluating}. Qwen3.5-9B improves no-target rejection over the 4B model on both datasets, although a gap to single-target performance remains. Together, these results motivate combining the comparatively low instruction sensitivity of the evaluated VLMs with dedicated spatial prediction.

\subsection{RQ2: Grounding Performance}
\label{sec:rq2}

We evaluate whether VGE improves grounding across instruction variants and varying numbers of targets on gRefCOCO, RoboRefIt and Ref-ZOM. We compare VGE with task-trained grounding experts, RECANTFormer, and zero-shot and LoRA-adapted Qwen3.5-9B and Groma. All task-specific training uses each dataset’s original instructions.

As shown in Table~\ref{tab:rq2}(a), VGE achieves the highest context-augmented F1 across all five test splits, with a difference of at most 1.2 percentage points between instruction conditions within each split. Grounding DINO and LLMDet remain sensitive to added context even after the task-specific training. Qwen3.5-9B with LoRA performs better on original and paraphrased instructions on gRefCOCO Test B and Ref-ZOM, but struggles under context augmentation. On Ref-ZOM, for example, LoRA achieves 90.7\% F1 on original instructions and 81.7\% with added context, whereas VGE achieves 86.1\% under both conditions. These results indicate that VGE maintains competitive grounding performance under added context, supporting its use in settings where robotics manipulation instructions are embedded in broader conversational context~\cite{moskalenko2026bring}.

As shown in Table~\ref{tab:rq2}(b), the breakdown by number of targets shows the greatest improvements on robotic scenes. VGE achieves the highest F1 in all three target groups on both RoboRefIt splits. For multiple-target references, it reaches 96.0\% and 81.3\% on Test A and Test B, compared with 90.7\% and 76.3\% for LoRA-adapted Qwen3.5-9B. On the general scenes, VGE improves multiple-target recovery on both gRefCOCO splits but obtains lower no-target accuracy. On Ref-ZOM, LoRA performs better on no-target and multiple-target references. Thus, strong context-augmented performance does not imply uniformly better results across target counts; both aspects require evaluation.

\begin{figure}[!htb]
\centering
\includegraphics[width=\linewidth]{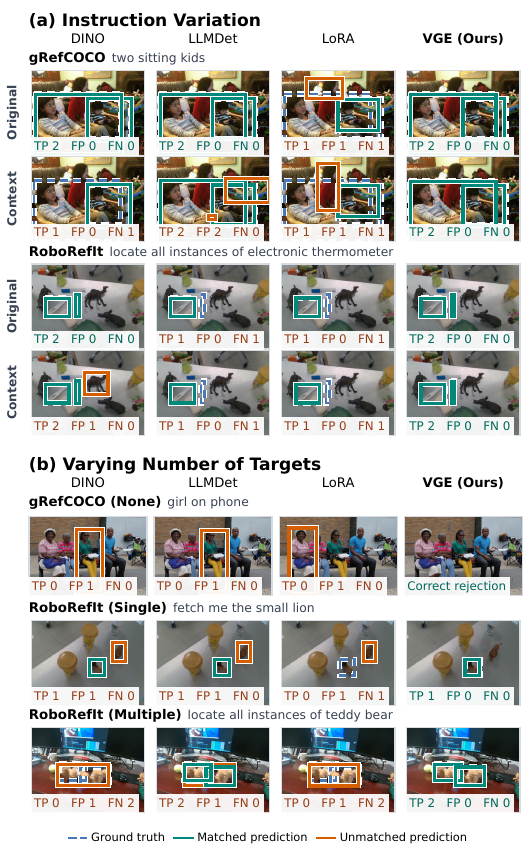}
\vspace{-15px}
\caption{\textbf{Qualitative grounding comparisons.} (a) Predictions under original and context-augmented instructions with unchanged targets. (b) Examples with no, one and multiple targets. Dashed blue boxes mark ground truth; solid teal and orange boxes mark matched and unmatched predictions, respectively. TP/FP/FN: true positives, false positives and false negatives. DINO denotes Grounding DINO; LoRA denotes the Qwen3.5-9B baseline with LoRA adaptation.}
\label{fig:rq2}
\end{figure}

Figure~\ref{fig:rq2} illustrates these differences through individual predictions. In panel~(a), added context causes Grounding DINO to miss one of the two sitting children and select an irrelevant object in the thermometer example. VGE recovers both requested instances under each instruction condition. The LoRA baseline, by contrast, retains the same TP/FP/FN counts across variants while continuing to miss a target in both examples, illustrating that stable performance need not imply correct grounding. Panel~(b) shows complementary target-selection errors. VGE returns an empty set for the absent target, selects only the requested small lion and separates the two teddy bears, whereas the baselines produce extra predictions or miss requested instances. These examples show how instruction sensitivity and target-selection errors affect predictions.

\subsection{RQ3: Is Expert Training Sufficient?}
\label{sec:rq3}

\begin{table}[!htb]
\centering
\caption{\textbf{Effect of VLM adaptation on grounding.} Mean sample F1 (\%), averaged equally over original, paraphrased and context-augmented instructions.}
\label{tab}
\footnotesize
\setlength{\tabcolsep}{2pt}
\renewcommand{\arraystretch}{1.1}
\begin{tabular}{@{}lcccc@{}}
\toprule
& \multicolumn{2}{c}{gRefCOCO}
& \multicolumn{2}{c}{RoboRefIt} \\
\cmidrule(lr){2-3}\cmidrule(lr){4-5}
Setting & Test A & Test B & Test A & Test B \\
\midrule
Frozen
& $79.98_{\pm 0.41}$ & $72.52_{\pm 0.55}$
& $95.02_{\pm 0.15}$ & $89.67_{\pm 0.26}$ \\
Expert-gradient
& $80.55_{\pm 0.40}$ & $73.64_{\pm 0.54}$
& $\mathbf{95.17}_{\pm 0.15}$ & $89.27_{\pm 0.27}$ \\
CE-adapted
& $\mathbf{81.95}_{\pm 0.38}$ & $\mathbf{77.14}_{\pm 0.50}$
& $94.20_{\pm 0.15}$ & $89.16_{\pm 0.26}$ \\
Adapted-frozen
& $\mathbf{81.95}_{\pm 0.38}$ & $75.78_{\pm 0.50}$
& $94.97_{\pm 0.15}$ & $\mathbf{89.96}_{\pm 0.25}$ \\
\bottomrule
\end{tabular}
\end{table}

In this experiment, we examine whether training only the grounding expert and both interfaces is sufficient to achieve performance comparable to configurations that also adapt the VLM. All configurations use original training instructions and train the expert and both interfaces. \emph{Frozen}, our main configuration, keeps the original VLM fixed. \emph{Expert-gradient} additionally updates rank-8 VLM LoRA parameters through the expert loss. \emph{CE-adapted} instead updates these parameters through coordinate-generation cross-entropy, with expert-loss gradients stopped at the VLM interface. \emph{Adapted-frozen} uses the LoRA adapter trained for the RQ2 baseline and freezes the adapted VLM throughout expert training. Implementation details are provided in Supplementary \cref{sec:supp_comparisons}.

Table~\ref{tab} shows that expert-and-interface training alone achieves performance close to all tested VLM adaptation strategies on RoboRefIt. Frozen remains within 0.82 percentage points of every alternative, and no adaptation strategy improves both test splits. This conclusion does not extend uniformly to gRefCOCO, where all three adaptation strategies outperform Frozen on both splits. However, continued VLM updates during expert training are not required to obtain much of this improvement. Adapted-frozen matches CE-adapted on Test A and achieves 75.78\% F1 on Test B, compared with 72.52\% for Frozen and 77.14\% for CE-adapted. Thus, training only the expert and interfaces is sufficient to approach the tested alternatives in most cases.

The limitation of keeping the original VLM frozen is most evident on gRefCOCO Test B, where CE-adapted improves F1 by 4.62 percentage points. These splits inherit RefCOCO's distinct image partitions, with Test A focusing on multiple people and Test B on multiple instances of non-person object categories~\cite{he2023grec}. Previous GREC evaluations also report lower performance on Test B across several grounding models~\cite{ding2026grex}. These observations show that Expert-and-interface training of the VGE is competitive in most settings, while additional VLM adaptation remains valuable for the broader range of object categories. Overall, training only the grounding expert and interfaces provides a practical balance between grounding performance and training cost by avoiding VLM parameter updates, while task-specific VLM adaptation remains beneficial in some settings, particularly on gRefCOCO Test B.

\subsection{RQ4: Inference Latency}
\label{sec:rq4}
We compare the inference latency of VGE, Grounding DINO and LoRA-adapted Qwen3.5-9B using the same hardware. For each dataset, we sample 300 references and evaluate their original, paraphrased and context-augmented instructions. Each method processes identical inputs in three rounds, yielding 2,700 measurements per dataset. Latency excludes image and text preprocessing.

\begin{table}[t]
\centering
\caption{\textbf{Inference latency.} Values are in milliseconds, measured on identical inputs and hardware, excluding image and text preprocessing. P50/P95: median/95th-percentile latency.}
\label{tab:latency}
\setlength{\tabcolsep}{3pt}
\renewcommand{\arraystretch}{1.15}
\begin{tabular}{@{}llrrr@{}}
\toprule
Dataset & Method & Mean & P50 & P95 \\
\midrule
\multirow{3}{*}{gRefCOCO} & Grounding DINO & 68.1 & 67.0 & 72.3 \\
 & Qwen3.5-9B(LoRA) & 1820.8 & 728.4 & 2561.5 \\
 & VGE & 139.9 & 135.4 & 174.3 \\
\midrule
\multirow{3}{*}{RoboRefIt} & Grounding DINO & 66.4 & 65.5 & 69.1 \\
 & Qwen3.5-9B(LoRA) & 737.5 & 736.0 & 1396.5 \\
 & VGE & 140.0 & 138.0 & 145.1 \\
\midrule
\multirow{3}{*}{Ref-ZOM} & Grounding DINO & 70.1 & 68.7 & 75.8 \\
 & Qwen3.5-9B(LoRA) & 872.1 & 738.9 & 1454.9 \\
 & VGE & 145.1 & 140.7 & 178.6 \\
\bottomrule
\end{tabular}
\end{table}

Table~\ref{tab:latency} shows that VGE achieves a mean latency of 139.9--145.1 ms, yielding $13.0\times$, $5.3\times$ and $6.0\times$ speedups over LoRA-adapted Qwen3.5-9B on gRefCOCO, RoboRefIt and Ref-ZOM, respectively. VGE also reduces median and 95th-percentile latency across all three datasets, while requiring approximately $2.1\times$ the mean latency of Grounding DINO. Thus, VGE incurs additional computation to incorporate VLM context but remains substantially faster than the VLM baselines with autoregressive grounding. Together with the VLM adaptation results, these findings illustrate the practical trade-offs of using a frozen VLM for contextual guidance while relying on a specialised expert for direct spatial prediction.

\subsection{RQ5: Interface Contributions}
\label{sec:rq5}
We evaluate the individual and combined contributions of layerwise joint attention and text-conditioned instance scoring on gRefCOCO and RoboRefIt using four configurations. \emph{DINO-only} uses the grounding expert without either interface. \emph{Joint-attn-only} adds only layerwise joint attention, while \emph{Scoring-only} adds only text-conditioned instance scoring. VGE combines both interfaces. 

\begin{table}[!htb]
\centering
\caption{\textbf{Contributions of the two interfaces.} Mean sample F1 (\%), averaged equally over the three instruction conditions. }
\vspace{-5px}
\label{tab:rq5}
\footnotesize
\setlength{\tabcolsep}{2pt}
\renewcommand{\arraystretch}{1.1}
\begin{tabular}{@{}lcccc@{}}
\toprule
& \multicolumn{2}{c}{gRefCOCO} & \multicolumn{2}{c}{RoboRefIt} \\
\cmidrule(lr){2-3}\cmidrule(lr){4-5}
Setting & Test A & Test B & Test A & Test B \\
\midrule
DINO-only & $69.07_{\pm 0.45}$ & $60.39_{\pm 0.57}$ & $82.25_{\pm 0.17}$ & $76.29_{\pm 0.26}$ \\
Joint-attn-only & $76.70_{\pm 0.39}$ & $71.49_{\pm 0.49}$ & $79.66_{\pm 0.16}$ & $73.47_{\pm 0.25}$ \\
Scoring-only & $78.51_{\pm 0.40}$ & $72.91_{\pm 0.52}$ & $83.61_{\pm 0.15}$ & $78.29_{\pm 0.24}$ \\
VGE & $79.98_{\pm 0.41}$ & $72.52_{\pm 0.55}$ & $95.02_{\pm 0.15}$ & $89.67_{\pm 0.26}$ \\
\bottomrule
\end{tabular}
\end{table}

Table~\ref{tab:rq5} shows that Scoring-only improves F1 over DINO-only on all four splits. Joint-attn-only improves F1 on gRefCOCO but reduces it on RoboRefIt. Combining both interfaces yields the highest F1 on three of the four splits. On RoboRefIt, VGE exceeds Scoring-only by 11.41 and 11.38 percentage points on Test A and Test B, respectively. On gRefCOCO, the corresponding differences are smaller, with a gain of 1.47 points on Test A and a reduction of 0.39 points on Test B. These results show that instance scoring provides consistent improvements across the evaluated splits, whereas the contribution of joint attention depends on the dataset and its interaction with scoring. Their combination is particularly beneficial on RoboRefIt, where neither interface alone approaches the performance of VGE.

This interaction suggests that contextual query refinement is most effective when accompanied by instruction-aware candidate scoring. Joint attention changes the representations used for box prediction and native matching but may not benefit localisation equally. Text-conditioned instance scoring may help align candidate selection with guidance from the VLM's instruction understanding, a benefit attributable to task-relevant attention guidance~\cite{zhu2026lesionattn}. Such coordination is important because accurate localisation alone does not ensure that matching objects receive appropriate scores~\cite{feng2021tood}. This offers a possible explanation for the gains obtained by combining both interfaces.

\section{Conclusion}
\label{sec:conclusion}

We introduced VGE, which integrates a frozen VLM with a trainable grounding expert through two interfaces. VGE uses multimodal context to guide object-query refinement and target selection while predicting bounding boxes directly. Across three datasets, it achieves the highest context-augmented F1 among the evaluated methods, with at most 1.2 percentage points of variation across instruction conditions within each split. On RoboRefIt, it also leads across no-target, single-target and multiple-target cases, supporting its effectiveness under both instruction variation and varying numbers of targets. VGE achieves these results with a $5.3$--$13.0\times$ inference speedup over the evaluated VLM baseline. 
A limitation of VGE is that freezing the VLM can leave room for further gains through task-specific adaptation. Nevertheless, results across two datasets support training only the grounding expert and interfaces as a competitive default that avoids VLM parameter updates.
Overall, VGE offers an effective approach to grounding under instruction variation and varying numbers of targets, with particularly promising grounding results in robotic scenes.

\section*{LLM Usage Statement}
OpenAI ChatGPT and Codex assisted with manuscript drafting and editing, LaTeX preparation and formatting, and literature discovery. Codex also assisted with code development. We used \texttt{qwen3.8-flash} to generate the paraphrased and context-augmented instruction variants used for evaluation. The original dataset images and original instructions were left unchanged. The VLMs used as model components and experimental baselines are described in the experimental setup. The authors retain responsibility for the content and results of this work.

{
\small
\bibliographystyle{ieeenat_fullname}
\bibliography{main}
}

\clearpage
\setcounter{page}{1}
\renewcommand{\thepage}{S\arabic{page}}
\maketitlesupplementary
\setcounter{section}{0}
\renewcommand{\thesection}{S\arabic{section}}
\renewcommand{\theHsection}{supp.\arabic{section}}
\setcounter{table}{0}
\renewcommand{\thetable}{S\arabic{table}}
\renewcommand{\theHtable}{supp.\arabic{table}}
\setcounter{figure}{0}
\renewcommand{\thefigure}{S\arabic{figure}}
\renewcommand{\theHfigure}{supp.\arabic{figure}}
\setcounter{equation}{0}
\renewcommand{\theequation}{S\arabic{equation}}
\renewcommand{\theHequation}{supp.\arabic{equation}}

\section{Authors}
\label{sec:authors}

\begin{center}
\small
\setlength{\tabcolsep}{4pt}
\renewcommand{\arraystretch}{1.3}
\begin{tabular}{@{}lll@{}}
Mingcheng Zhu\textsuperscript{1,2,*} &
Wentian Xu\textsuperscript{1,2,*} &
Tao Lin\textsuperscript{2,3,*} \\
Shiying Duan\textsuperscript{2} &
Ziyun Liang\textsuperscript{1,2} &
Xuecheng Qi\textsuperscript{2} \\
Junyang Chen\textsuperscript{2} &
Kuo Li\textsuperscript{2} &
Zhifan Wan\textsuperscript{2} \\
Dong Wang\textsuperscript{2} &
Richard Wang\textsuperscript{2} &
Yang Gao\textsuperscript{2,4,$\dagger$}
\end{tabular}

\medskip
{\footnotesize
\textsuperscript{1}University of Oxford \quad
\textsuperscript{2}Spirit AI\\[2pt]
\textsuperscript{3}University of Chinese Academy of Sciences\\[2pt]
\textsuperscript{4}Tsinghua University
\par}

\medskip
{\footnotesize \textsuperscript{*}Equal contribution. \textsuperscript{$\dagger$}Corresponding author.}
\end{center}

\section{Datasets}
\label{sec:supp_setup}
We evaluate generalised referring expression comprehension on three datasets. gRefCOCO~\cite{liu2023gres,he2023grec} and Ref-ZOM~\cite{Hu_2023_ICCV} cover general scenes, while RoboRefIt~\cite{lu2023vlgrasp} covers robotic scenes. Each reference pairs an instruction with an image and an annotated set of target boxes. All three datasets include no-target, single-target and multiple-target references. We retain the original images, instructions and target annotations, including the annotated no-target and multiple-target examples of RoboRefIt.

We retain the original test partitions. Where no validation split is provided, we reserve 1,200 proportionally sampled training references for validation. Table~\ref{tab:supp_datasets} summarises the resulting splits. Counts refer to distinct references before instruction augmentation; the three instruction variants do not triple the reference count.

\begin{table}[!htb]
\centering
\caption{\textbf{Dataset statistics.} Unique images and references in our experimental splits. None, Single and Multiple count references with zero, one and multiple ground-truth targets, respectively. }
\label{tab:supp_datasets}
\scriptsize
\setlength{\tabcolsep}{2.4pt}
\renewcommand{\arraystretch}{1.15}
\begin{tabular}{@{}llrrrrr@{}}
\toprule
Dataset & Split & Images & Refs. & None & Single & Multiple \\
\midrule
gRefCOCO & Train & 17,709 & 81,966 & 12,564 & 42,004 & 27,398 \\
 & Validation & 785 & 1,200 & 400 & 400 & 400 \\
 & Test A & 750 & 6,265 & 749 & 1,975 & 3,541 \\
 & Test B & 750 & 4,618 & 750 & 1,807 & 2,061 \\
\midrule
RoboRefIt & Train & 7,529 & 69,820 & 7,529 & 54,351 & 7,940 \\
 & Validation & 400 & 1,200 & 400 & 400 & 400 \\
 & Test A & 1,859 & 16,636 & 1,859 & 12,767 & 2,010 \\
 & Test B & 1,083 & 10,575 & 1,083 & 8,522 & 970 \\
\midrule
Ref-ZOM & Train & 42,551 & 66,416 & 9,210 & 44,211 & 12,995 \\
 & Validation & 1,194 & 1,200 & 400 & 400 & 400 \\
 & Test & 11,329 & 21,770 & 2,327 & 12,056 & 7,387 \\
\bottomrule
\end{tabular}
\end{table}

All task-specific training uses original instructions. For evaluation, each reference is paired with a paraphrase and a context-augmented variant. Paraphrases reword the request; context augmentation adds background information without changing the target requirements. Both retain the original image and annotated target set.

\section{Implementation Details}
\label{sec:supp_implementation}
\subsection{Interface Implementation and Training}
\label{sec:supp_interfaces}

VGE pairs the grounding expert's six decoder layers, in depth order, with the VLM's last six full-attention layers, excluding linear-attention and sliding-window-attention layers. At each paired VLM layer, keys are extracted after native key normalisation and positional transformations, and values are taken from the native value projection. The trainable mappings $f_K^\ell$ and $f_V^\ell$ concatenate features across VLM heads, project them to the expert's attention dimension and reshape them into expert heads. Joint attention includes all non-padding image, text and special-token positions. The adapted VLM keys and values share a single attention softmax and the expert's native output projection with the expert keys and values. Each decoder layer uses an independent, unconstrained scalar gate $\alpha_\ell$, initialised to zero. The gated combination is applied after output projection and replaces the native self-attention output, retaining the subsequent residual, normalisation and cross-attention operations.

For text-conditioned instance scoring, $H_P$ contains features from the final VLM block before output normalisation at non-padding text positions, including template and system tokens where present. We apply layer normalisation followed by the linear mapping $f_H$, which projects these features to 256 dimensions with a bias and no activation. Each layer-normalised final object query $q_i$ attends to the projected features through a trainable \texttt{nn.MultiheadAttention} module with eight 32-dimensional heads, biased projections and zero dropout. Concatenating $q_i$ with the resulting context $c_i$ gives a 512-dimensional vector. The linear mapping $f_s$ converts this vector into a scalar correction to the native matching logit. Its weights and bias are initialised to zero, preserving the expert's native scores at initialisation.

The main configuration uses the public \texttt{Qwen/Qwen3.5-9B} and \texttt{IDEA-Research/grounding-dino-base} checkpoints. We freeze the VLM and train the complete grounding expert and both interfaces on the original instructions using AdamW~\cite{loshchilov2019adamw}. Peak learning rates are $10^{-4}$ for the expert and $5\times10^{-4}$ for the interface parameters. We use weight decay 0.01, $\beta=(0.9,0.999)$ and $\epsilon=10^{-8}$. Training uses a global batch size of 32. Linear warmup covers the first 10\% of optimisation steps, followed by cosine decay to 10\% of each peak learning rate. The global gradient norm is clipped to 1.0. Frozen VLM weights are stored in BF16, mixed-precision computation uses BF16 autocast, and expert master parameters remain in FP32.

Training uses the final-layer objective in Equation~\ref{eq:loss} with $\lambda_1=5$ and $\lambda_2=2$, following the setting in the Grounding DINO~\cite{liu2024groundingdino}. Focal loss uses $\alpha=0.25$ and $\gamma=2$, with unmatched queries assigned negative classification targets. We select the checkpoint with the highest mean sample F1 on original validation instructions. Within each dataset, the same checkpoint is used for Test A and Test B.

\subsection{Input Preprocessing}
\label{sec:supp_preprocessing}

The grounding expert and VLM process the same image and instruction using their respective preprocessing pipelines. For Grounding DINO, RGB images are resized while preserving the aspect ratio, with a target shorter edge of 800 pixels and a maximum longer edge of 1,333 pixels. Images are then padded within each batch. Pixel values are rescaled by $1/255$ and normalised using mean $(0.485,0.456,0.406)$ and standard deviation $(0.229,0.224,0.225)$. Qwen uses its native image processor with minimum and maximum pixel counts of 65,536 and 1,048,576, respectively. We apply no random flips, crops or colour augmentation.

The grounding expert accepts up to 512 text tokens, with longer sequences truncated from the left. The VLM input is limited to 8,192 tokens, including image and template tokens. The main configuration uses no system prompt and disables thinking mode. Training uses original instructions, while evaluation uses the original, paraphrased or context-augmented instruction specified by the evaluation condition. Assistant target outputs are excluded from the inputs used to compute expert predictions.

\subsection{Baseline Configurations and Inference}
\label{sec:supp_comparisons}

We evaluate Grounding DINO-base and LLMDet-base using both public checkpoints and task-specific fine-tuning. Grounding DINO and the Hugging Face implementation of LLMDet score candidates using the maximum token-matching probability. The native LLMDet implementation used for task-specific fine-tuning instead averages these probabilities over the referring expression. Each detector configuration is calibrated independently on original validation instructions.

The Qwen LoRA baseline uses coordinate-generation cross-entropy with all-linear adapters of rank 8, a scaling factor of 16, zero dropout and a learning rate of $5\times10^{-5}$. The base VLM parameters remain frozen. VGE's adaptation variants use the same LoRA settings. Expert-gradient updates the adapters through the expert loss, whereas CE-adapted uses coordinate-generation cross-entropy with weight 1.0 and blocks expert-loss gradients into the VLM. Adapted-frozen keeps a previously trained LoRA adapter fixed during expert training. The grounding expert and both interfaces remain trainable in all VGE configurations.

Confidence thresholds are selected to maximise mean sample F1 on original validation instructions and remain fixed across test splits, instruction variants and target counts within each dataset. VGE and task-trained Grounding DINO search thresholds range from 0.05 to 0.95 in increments of 0.05, while native LLMDet uses a predefined grid between 0.01 and 0.90. The selected thresholds on gRefCOCO, RoboRefIt and Ref-ZOM are $(0.40,0.25,0.30)$ for VGE, $(0.45,0.35,0.30)$ for Grounding DINO and $(0.40,0.25,0.30)$ for LLMDet. Candidates meeting the threshold undergo hard non-maximum suppression at IoU 0.5, with at most 900 output boxes. 

Zero-shot Qwen and its LoRA baseline generate up to 1,024 tokens and share the same JSON parser. Outputs contain XYXY bounding boxes with coordinates normalised to $[0,1000]$, or \texttt{[]} when no target is present. The parser accepts a single box, a \texttt{bbox\_2d} object or a list of such objects, optionally enclosed in a Markdown code fence. Every box must contain finite coordinates satisfying $0\leq x_1<x_2\leq1000$ and $0\leq y_1<y_2\leq1000$. Malformed outputs, invalid boxes and generations truncated by the token limit are treated as failed predictions. Valid boxes are converted to pixel coordinates without further filtering or non-maximum suppression.

\subsection{Evaluation Protocol}
\label{sec:supp_evaluation}

All methods are evaluated on the same references within each test split. Failed outputs remain included and receive a score of zero. We evaluate original, paraphrased and context-augmented instructions separately. Combined results weight the three instruction conditions equally, and results by target count use the corresponding no-target, single-target or multiple-target references.

Predicted and ground-truth boxes are matched greedily at an IoU threshold of 0.5. At each step, we select the remaining pair with the highest IoU and remove both boxes from further matching. Matched pairs are true positives ($TP$), unmatched predictions are false positives ($FP$), and unmatched targets are false negatives ($FN$). For references with at least one predicted or ground-truth box, sample F1 is
\begin{equation}
    F_1=\frac{2TP}{2TP+FP+FN}.
\end{equation}
A valid empty prediction receives a score of one when the ground-truth set is empty. Malformed or truncated outputs receive zero regardless of the target count. Mean sample F1 averages these scores across references. Exact-set accuracy requires all ground-truth targets to be matched without additional predictions. On no-target references, mean sample F1 equals no-target accuracy.

Reported standard deviations are calculated from 1,000 paired bootstrap resamples of reference IDs using seed~0. Each resample draws references with replacement and uses the same sampled IDs across methods and instruction conditions. References sharing an image are sampled individually. Model predictions and confidence thresholds remain fixed, and each metric is recomputed for every resample. We report scores on the complete evaluation set alongside bootstrap sample standard deviations, with error bars representing one standard deviation. Each configuration is trained once with seed~0, so these standard deviations describe reference-level resampling variability rather than variation across training runs.

\section{VLM and Grounding-Expert Choices}
\label{sec:supp_components}

\begin{figure}[!htb]
    \centering
    \includegraphics[width=0.98\linewidth]{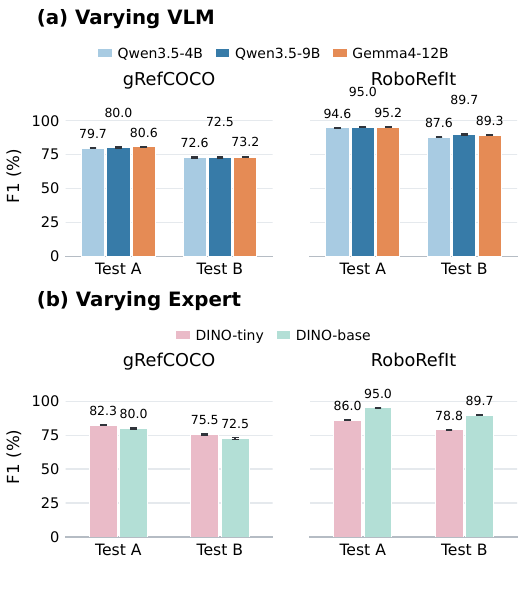}
    \caption{\textbf{Effect of VLM and grounding-expert choices.} Mean sample F1 (\%), averaged equally over original, paraphrased and context-augmented instructions. (a) Frozen VLM variants with Grounding DINO-base fixed. (b) Grounding-expert variants with frozen Qwen3.5-9B fixed. Error bars show standard deviations from 1,000 paired reference-bootstrap resamples.}
    \label{fig:scale}
\end{figure}

In this experiment, we examine how VLM and grounding expert choices affect the grounding performance of VGE. Figure~\ref{fig:scale}(a) compares frozen Qwen3.5-4B, Qwen3.5-9B and Gemma4-12B with Grounding DINO-base fixed. Panel~(b) compares Grounding DINO-tiny and Grounding DINO-base with frozen Qwen3.5-9B fixed. For each configuration, we train the grounding expert and both interfaces on original instructions using the same training budget and learning rates. Each VLM retains its native visual preprocessing. We report mean sample F1 averaged equally over the three instruction conditions.

VLM choice has a comparatively small effect across the four splits. Replacing Qwen3.5-9B with Qwen3.5-4B changes F1 by at most 2.09 percentage points, while replacing it with Gemma4-12B changes F1 by at most 0.64 percentage points. These results show that VGE maintains similar performance with either of the two tested VLM families after training the expert and interfaces. The smaller Qwen3.5-4B also retains much of the performance of the main configuration.

Grounding-expert choice has a larger and domain-dependent effect. Relative to Grounding DINO-base, Grounding DINO-tiny improves F1 by 2.30--3.01 percentage points on gRefCOCO but reduces it by 9.00--10.87 points on RoboRefIt. The larger expert therefore performs better on robotic scenes but is not uniformly preferable across the evaluated datasets. Within the tested configurations, grounding performance is more sensitive to expert choice than to VLM choice.

\end{document}